\PassOptionsToPackage{dvipsnames,svgnames}{xcolor}
\documentclass[preprint,12pt,authoryear]{elsarticle}
\usepackage[utf8]{inputenc}
\usepackage[T1]{fontenc}
\usepackage{lineno}
\usepackage{xcolor}
\usepackage{enumitem}
\usepackage{titlesec}
\usepackage{makecell}
\usepackage{multirow} 
\usepackage{multicol}
\usepackage{hyperref}
\usepackage{tabularx}
\usepackage{graphicx}
\usepackage{booktabs}
\usepackage{caption}

\usepackage{amssymb}
\usepackage{amsmath}

\journal{Artificial Intelligence in Agriculture}

\begin{document}

\begin{frontmatter}

\title{AgriCHN: A Comprehensive Cross-domain Resource for Chinese Agricultural Named Entity Recognition}

\author[label1]{Lingxiao Zeng\fnref{equal}}
\author[label1]{Yiqi Tong\fnref{equal}\corref{corresponding}}
\author[label2]{Wei Guo}
\author[3,4,5]{Huarui Wu}
\author[label1]{Lihao Ge}
\author[6]{Yijun Ye}
\author[label2]{Fuzhen Zhuang}
\author[label1]{Deqing Wang}
\author[3,4,5]{Wei Guo\corref{corresponding}}
\author[3,4,5]{Cheng Chen}

\cortext[corresponding]{Corresponding author.}
\fntext[equal]{The two authors contribute equally to this work.}

\affiliation[label1]{organization={School of Computer Science and Engineering, Beihang University},
            city={Beijing},
            postcode={100191},
            country={China}}

\affiliation[label2]{organization={School of Artificial Intelligence, Beihang University},
            city={Beijing},
            postcode={100191},
            country={China}}
            
\affiliation[3]{organization={Beijing Research Center for Information Technology in Agriculture, Beijing Academy of Agriculture and Forestry Sciences},
            city={Beijing},
            postcode={100097},
            country={China}}
            
\affiliation[4]{organization={Laboratory of Digital Village Technology, Ministry of Agriculture and Rural Affairs},
            city={Beijing},
            postcode={100097},
            country={China}}

\affiliation[5]{organization={National Engineering Research Center for Information Technology in Agriculture},
            city={Beijing},
            postcode={100097},
            country={China}}
            
\affiliation[6]{organization={Laboratory of Multimodal Artificial Intelligence Systems (MAIS), Institute of Automation, Chinese Academy of Sciences},
            city={Beijing},
            postcode={100190},
            country={China}}

\begin{abstract}
Agricultural named entity recognition is a specialized task focusing on identifying distinct agricultural entities within vast bodies of text, including crops, diseases, pests, and fertilizers. It plays a crucial role in enhancing information extraction from extensive agricultural text resources. However, the scarcity of high-quality agricultural datasets, particularly in Chinese, has resulted in suboptimal performance when employing mainstream methods for this purpose. Most earlier works only focus on annotating agricultural entities while overlook the profound correlation of agriculture with hydrology and meteorology. To fill this blank, we present AgriCHN, a comprehensive open-source Chinese resource designed to promote the accuracy of automated agricultural entity annotation. The AgriCHN dataset has been meticulously curated from a wealth of agricultural articles, comprising a total of 4,040 sentences and encapsulating 15,799 agricultural entity mentions spanning 27 diverse entity categories. Furthermore, it encompasses entities from hydrology to meteorology, thereby enriching the diversity of entities considered. Data validation reveals that, compared with relevant resources, AgriCHN demonstrates outstanding data quality, attributable to its richer agricultural entity types and more fine-grained entity divisions. A benchmark task has also been constructed using several state-of-the-art neural NER models. Extensive experimental results highlight the significant challenge posed by AgriCHN and its potential for further research.
\end{abstract}




\end{frontmatter}



\section{Introduction}
\begin{figure}[t]
\centering
\includegraphics[width=\linewidth]{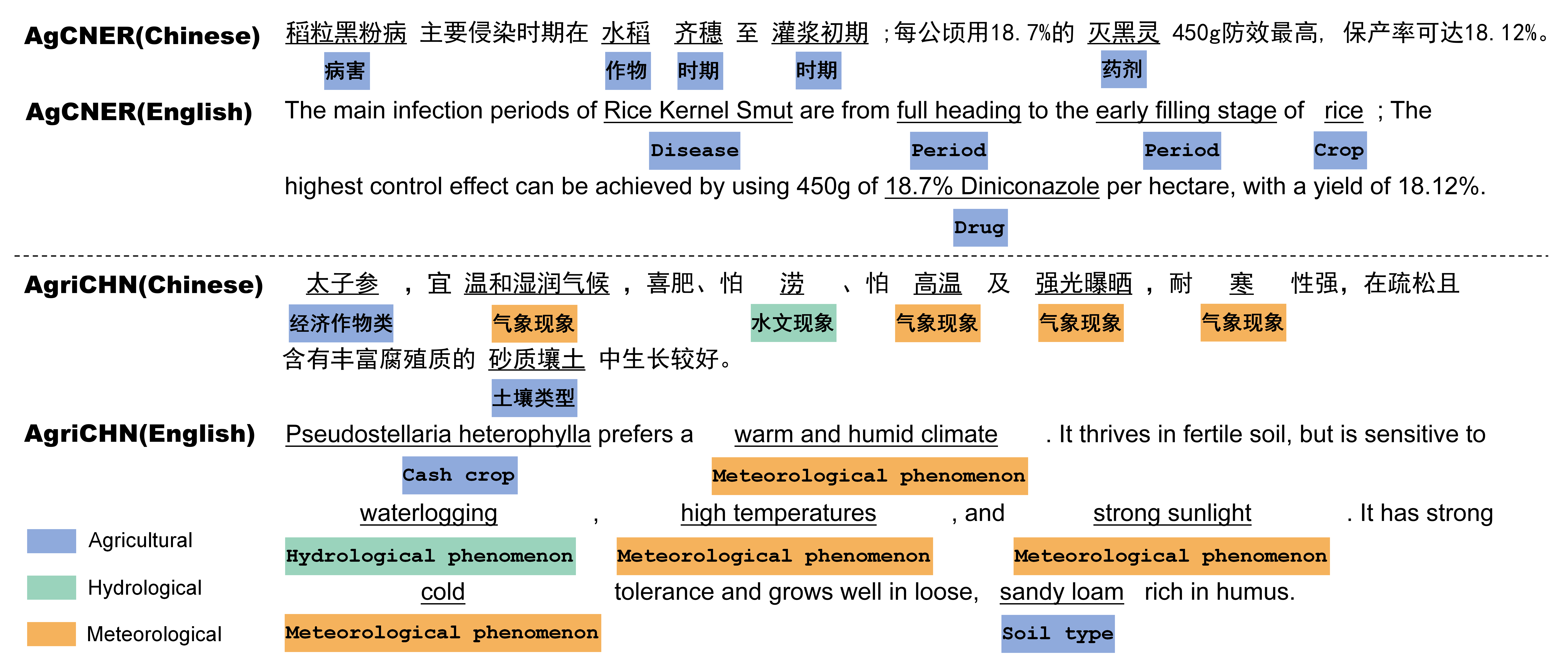}
\caption{Entity annotation examples from AgriCHN and AgCNER \citep{guochinese2020}, highlighting AgriCHN's cross-domain coverage versus AgCNER's agriculture-specific focus.}
\label{fig:sample_agri_NER}
\end{figure}

Chinese named entity recognition (NER) plays a crucial role in natural language processing (NLP). Its primary objective is to identify particular types of entities from provided texts. Compared to general Chinese NER that encompasses a wide range of entities, domain-specific Chinese NER is specialized in extracting entities within a specific knowledge domain \citep{Fang_2021_TEBNER}. As a branch of domain-specific Chinese NER, agricultural Chinese NER has become an increasingly popular research topic due to the growing value of agricultural entities. \hyperref[fig:sample_agri_NER]{Fig. \ref*{fig:sample_agri_NER}} shows samples of recognition of agricultural entities. These labeled entities, which contain abundant information about agriculture, are essential to building agricultural knowledge graphs \citep{nath2024bangladesh} and information systems \citep{mushi2024designing}.

From a broader perspective, accurately identifying agricultural entities lays a solid foundation for advancing in-depth agricultural data analysis, informing government policy decisions, and facilitating the development of intelligent agricultural infrastructure. However, compared to Chinese NER in other domains such as news \citep{tu2024named}, healthcare \citep{hu2024improving}, and education \citep{li2023eduner}, recognition of agricultural entities faces several unique challenges.

First, agricultural entities appear in diverse types of texts, ranging from scientific papers to news reports and policy documents. This diversity in text sources introduces varying terminology expressions and writing styles, making standardized entity recognition more complex. Second, the agricultural domain is characterized by entity variations, particularly in crop names and plant diseases, where multiple aliases are common. The complexity is further amplified by the hybrid nature of many agricultural entities, which often combine letters, numbers, Chinese characters, and punctuation marks, posing significant challenges for feature extraction. Third, agricultural NER must extend beyond pure agricultural terms to identify entities from closely related domains such as hydrology and meteorology, as these fields are inherently interconnected in practical agricultural production and daily operations.

To address these challenges, a comprehensive human-annotated NER dataset with agriculture-related entities is urgently required. Recent works have proposed available corpora for agricultural NER in Chinese and English, including AGRONER \citep{G._2023_AGRONER}, AgriNER \citep{de2023agriner}, AgCNER \citep{guochinese2020, yao2024agcner}, and CCNEDP \citep{Wang_2022_Named}.These corpora, derived from research papers, authoritative agricultural websites, official databases, and Wikipedia, have undergone rigorous manual annotations and screening to ensure data quality. However, two significant limitations persist in existing datasets. Current agricultural NER datasets typically contain fewer than 20 coarse-grained entity types, which constrains the extraction of refined information for downstream tasks. Moreover, these datasets focus solely on agricultural entities while overlooking the crucial interconnections with hydrology and meteorology, making them inadequate for practical agricultural applications.

The major contributions of this work are summarized as follows:
\begin{itemize}[]
\item We present AgriCHN\footnote{We make our proposed AgriCHN and other relevant documents publicly available at \href{http://github.com/SleeperZLX/AgriCHN-2023/tree/main}{http://github.com/SleeperZLX/AgriCHN-2023/tree/main}.}, a new open-source comprehensive human-annotated dataset for Chinese agricultural named entity recognition. Different from other datasets, AgriCHN is a cross-domain Chinese agricultural NER dataset that combines corresponding entities from hydrology and meteorology. Specifically, it contains 27 carefully designed entity types, e.g., more fine-grained crop types, water bodies, and meteorological phenomena. AgriCHN is composed of 4,040 sentences from full-text high-quality agricultural articles, paragraphs about smart agriculture without entities, and sentences extracted from other publicly available datasets.
\item We design a novel data pre-annotation process enhanced by Large Language Model (LLM). This approach utilizes the LLM with the standardized prompt to perform relation extraction on input sentences and achieve data filtering by indirectly assessing the richness of agricultural entities through the number of relations identified within the sentences. Additionally, we convert relations into entity annotations as preliminary labeling results, providing insights for the subsequent construction of NER datasets.
\item We set benchmarks for AgriCHN by exploiting the latest deep learning methods such as BiLSTM \citep{Huang_2015_Bidirectional}, step-wise attention \citep{vaswani2017attention}, and pre-trained language models (PLMs) \citep{devlin-etal-2019-bert}, etc. The experimental results demonstrate that the AgriCHN task deserves future research and that the AgriCHN dataset could be utilized as a reliable resource for verifying the accuracy of agricultural entity recognition.
\end{itemize}

\section{Materials and Methods}
\subsection{Overview}
The construction of the AgriCHN dataset follows rigorous labeling specifications through multiple iterations. Drawing from established practices \citep{malarkodi2016named,Liu_2022_novel,mol2024end}, we outline several key steps that are highly correlated with dataset quality:
\begin{itemize}[]
\item \textbf{Entity type selection.} Through iterative refinement, we establish a comprehensive set of agricultural entity types and their granularity levels, balancing practical utility with coverage scope.
\item \textbf{Document \& sentence selection and pre-annotation.} We employ the LLM to perform unrestricted relation extraction on agricultural articles, enabling effective filtering of entity-rich sentences and generating preliminary annotations.
\item \textbf{Annotation guidelines.} To ensure data consistency and quality, we develop detailed annotation guidelines for the annotators, including fundamental principles, task objectives, and general rules. We also provide additional explanations for special cases within AgriCHN.
\item \textbf{Human annotation.} We implement a double-blind annotation process with expert adjudication, where two annotators independently label the dataset based on pre-annotations and guidelines. Discrepancies are then resolved through review by a senior annotator, following standard inter-annotator agreement protocols.
\item \textbf{Dataset balancing.} To ensure comprehensive coverage, we augment the dataset with sentences from public sources to maintain adequate representation of general entities (e.g., Person, Time). Additionally, we incorporate texts without entities to better reflect real-world scenarios.
\end{itemize}

\subsection{Entity type selection}
Considering the lack of general NER datasets and entity classification standards in the agriculture domain \citep{Guo_2021_ACE-ADP}, our entity type selection is more based on actual needs: agricultural NER datasets should encompass a wide variety of comprehensive and fine-grained entity types due to the diverse range of agricultural entities and their intersections with other fields such as hydrology and meteorology. However, most previous datasets that contain a limited number of coarse-grained and overly fundamental types are not suitable for practical applications. Drawing inspiration from few-shot NER frameworks like Few-NERD \citep{Ding_2021_Few-NERD}, we develop our entity taxonomy through a systematic four-step process:

\textbf{Step 1:} We select several coarse-grained types as the primary version, including crop, livestock and poultry, aquatic entity, forestry entity, agricultural service, and other fundamental entities (person, organization, location, and time).

\textbf{Step 2:} We conduct an initial refinement of entity classifications within each coarse-grained category, guided by two core principles: (1) Agricultural core entities require fine-grained subdivisions to support downstream tasks such as knowledge graph construction. For instance, crops are further categorized into grains, beans, oil-seed crops, fruits, and vegetables. (2) For interdisciplinary domains, only entities that occur frequently in agricultural contexts merit inclusion. For example, while water bodies are commonly referenced in crop growth descriptions, hydrological equipment appears less frequently, leading to the inclusion of only ``water body" as an entity type. Based on these two principles, we summarize 54 potential entity types for the AgriCHN dataset.

\textbf{Step 3:} We conduct a pilot annotation phase to validate our entity classification scheme. Through this initial manual annotation process, several practical issues emerge: (1) certain entity types show extremely low frequency, warranting removal; (2) crop categorization based on characteristics (e.g., tea plants, fiber crops) is less effective than classification by productive use (e.g., cash crops, cereals); (3) non-agricultural plants and animals require consolidation under a single category named ``other organism".

\textbf{Step 4:} We refine our entity taxonomy based on these annotation insights. The resulting schema comprises 27 fine-grained entity types spanning agriculture, hydrology, and meteorology, as outlined in \hyperref[tab:details of entity types]{Table \ref*{tab:details of entity types}}.

\begin{table}[t]
\centering
\caption{\label{tab:details of entity types}Details of entity types in AgriCHN.}
\resizebox{\linewidth}{!}{%
\begin{tabular}{llcl|llcl}
\toprule
\#  & Tags (EN)      & Abbr.  & Type            & \#  & Tags (EN)      & Abbr.  & Type            \\ \midrule
1   & Cash crops      & CASH   & Agricultural    & 2   & Cereal crops      & CER    & Agricultural    \\
3   & Forage crops      & FORA   & Agricultural    & 4   & Vegetables          & VEG    & Agricultural    \\
5   & Fruits          & FRU    & Agricultural    & 6   & Aquatic products          & AQU    & Agricultural    \\
7   & Livestock          & LIV    & Agricultural    & 8   & Products          & PRO    & Agricultural    \\
9   & Other organisms        & OTHER  & Agricultural    & 10  & Pests, diseases, and weeds        & PDPW   & Agricultural    \\
11  & Fertilizers            & FER    & Agricultural    & 12  & Pesticides            & PES    & Agricultural    \\
13  & Nutrients        & NUT    & Agricultural    & 14  & Diseases            & DIS    & Agricultural    \\
15  & Agricultural technologies        & AGTE   & Agricultural    & 16  & Agricultural machinery        & AGEQ   & Agricultural    \\
17  & Agricultural infrastructure        & AGFA   & Agricultural    & 18  & Soil types        & SOIL   & Agricultural    \\
19  & Soil moisture        & SOMOI  & Agricultural    & 20  & Hydrological phenomena        & HYDPH  & Hydrological    \\
21  & Water bodies            & WATBO  & Hydrological    & 22  & Meteorological phenomena        & METPH  & Meteorological  \\
23  & Temperature            & TEMP   & Meteorological  & 24  & Location            & LOC    & General         \\
25  & Person            & PER    & General         & 26  & Time            & TIME   & General         \\
27  & Organization        & ORG    & General         & --  & --              & --     & --              \\ 
\bottomrule
\end{tabular}
}
\end{table}

\subsection{Document \& sentence selection and pre-annotation}
\label{sec:pre-annotation}
To construct a high-quality dataset with rich agricultural entities, we first collect documents from authoritative agricultural websites: China Agricultural Information Website\footnote{\href{http://www.agri.cn}{https://www.agri.cn}.}, China Crop Germplasm Information Website\footnote{\href{http://www.cgris.net}{http://www.cgris.net}.}, National Agricultural Mechanization Information Service Platform\footnote{\href{http://www.njztc.cn}{http://www.njztc.cn}.}, National Agricultural Technology Promotion Website\footnote{\href{https://www.natesc.org.cn}{https://www.natesc.org.cn}.}, and Agricultural Resource Website\footnote{\href{http://www.ampcn.com}{http://www.ampcn.com}.}.

Initially, we categorize the documents into seven groups: agricultural machinery (300 documents), farming guidance (200 documents), agricultural knowledge (300 documents), agricultural news (300 documents), plant diseases, pests and weeds (15 types, 1,270 documents), soil moisture (300 documents), and agricultural meteorology (300 documents). To optimize the composition of the dataset and balance the distribution of entities, we filter out documents with sparse agricultural knowledge. The final dataset comprises 200 documents on agricultural knowledge, 150 on pests and plant diseases, 100 on farming guidance, 50 on agricultural machinery, 30 on agricultural news, and 10 each for soil moisture and agricultural meteorology.

After document selection, it is extremely crucial to obtain valuable sentences from the documents so that the samples in AgriCHN contain sufficient agricultural entities and balanced in entity distribution. However, unlike previous work, which determined several principles to filter sentences \citep{Islamaj_2021_NLM-Chem,Zhang_2022_Research}, we introduce the LLM and relation extraction mechanism \citep{zaratiana2024autoregressive} to assist in our sentence selection. LLMs have demonstrated remarkable capabilities in NLP tasks, including high-accuracy entity recognition when provided with appropriate annotation prompts. Additionally, their API accessibility enables efficient task execution through simple request-response interactions. And the number of entity relationships is suitable for measuring text quality of sentence to a certain degree, better than artificial rules. Considering these advantages, we exploit relation extraction as an auxiliary task for choosing sentences and choose ChatGPT-turbo-3.5\footnote{\href{https://chatgpt.com/}{https://chatgpt.com.}} as the base LLM.

\begin{figure}[t]
\centering
\includegraphics[width=\linewidth]{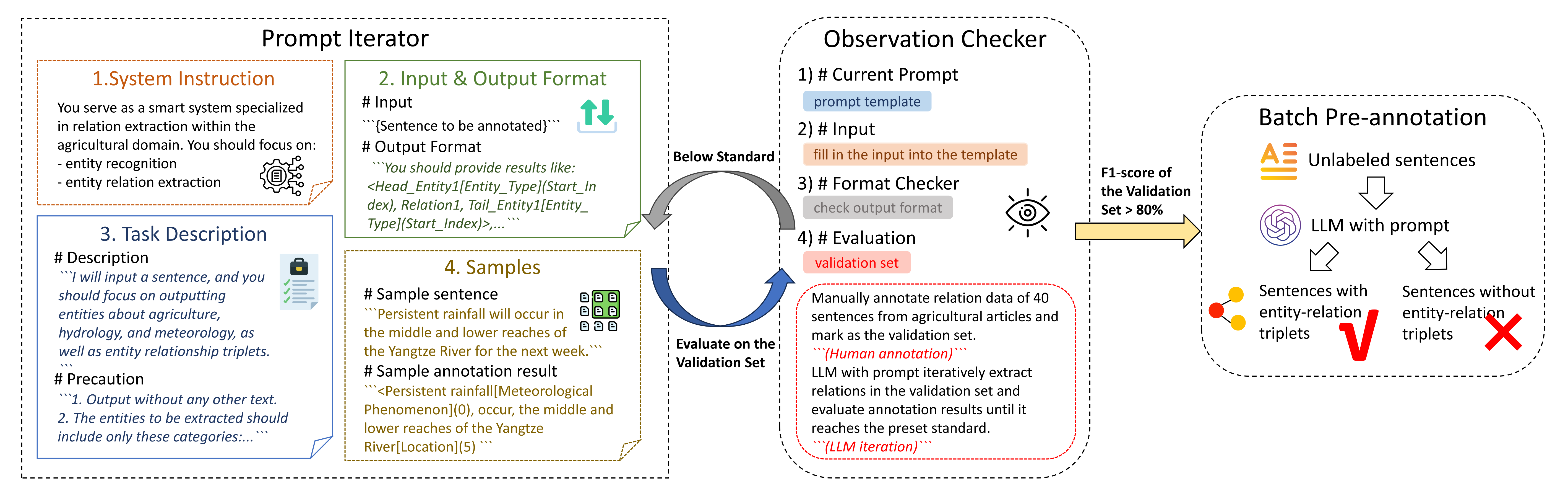}
\caption{Workflow of LLM-enhanced pre-annotation.}
\label{fig:LLM_Assist_Relation_Extraction}
\end{figure}

Specifically, we first perform preliminary screening and cleaning of sentences from each article. This involves removing redundant characters such as spaces, tabs, and newlines. Moreover, we filter out sentences that are either too long (exceeding 128 tokens) or too short (fewer than 5 characters) after cleaning. Subsequently, we implement LLM-enhanced pre-annotation for candidate sentences, with its overall workflow illustrated in \hyperref[fig:LLM_Assist_Relation_Extraction]{Fig. \ref*{fig:LLM_Assist_Relation_Extraction}}:

\begin{itemize}[]
\item \textbf{System instruction.} We initialize system instruction of the LLM to function as an intelligent relationship extraction system that identifies and extracts specific relationships from agricultural texts.
\item \textbf{Format definition.} We define standardized input and output formats for the LLM. Given an agricultural sentence $s$ as input, the output is a set of relation triplets $R = {r_1, r_2, ..., r_n}$, where each triplet $r_i$ is formulated as $(h, rel, t)$. Here, $h = (e_h, type_h, pos_h)$ and $t = (e_t, type_t, pos_t)$ represent the head and tail entities respectively, with $e$ denoting the entity text, $type$ indicating entity category, and $pos$ marking the starting position in sentence $s$.
\item \textbf{Task description and exemplification.} We show the LLM with task descriptions and key considerations. For example, extracted entities must belong to the categories in \hyperref[tab:details of entity types]{Table \ref*{tab:details of entity types}}. Meanwhile, we provide annotation examples for the LLM, which include detailed examples illustrating the desired extraction format and entity categories.
\item \textbf{Observation checker.} We implement an observation checker serving dual purposes. Primarily, it validates output conformity to the required format, eliminating duplicated data, non-standard outputs, and ambiguous information. Additionally, it facilitates iterative refinement of annotation results. We manually annotate 30–50 agricultural sentences as a preliminary validation set, enabling the checker to iteratively optimize the LLM prompt based on annotation performance. This optimization continues until achieving an annotation accuracy threshold of 80\%, ensuring reliability before scaling to large-scale pre-annotation.
\item \textbf{Pre-annotation and filtering.} Upon obtaining the optimal LLM prompt, we conduct batch pre-annotation while automatically filtering out sentences lacking entity relationships, thereby ensuring the quality and relevance of our dataset.
\end{itemize}

Through these processes, sentences for AgriCHN undergo refinement and pre-annotation. Given that the relation dataset already contains certain information pertaining to agricultural entities, we transform it into the initial version of AgriCHN in BIO format, thereby reducing a portion of our labor costs.

\subsection{Annotation guidelines}
Golden annotation guidelines are crucial for ensuring dataset consistency and reliability \citep{benikova2014nosta}. As a result, we expect annotation guidelines to become comprehensive and cover largest possible number of cases so that our annotators can avoid labeling errors. Meanwhile, the complete annotation guidelines for AgriCHN are publicly available to offer references for the development and advancement of agricultural NER tools. In this section, our annotation guidelines are introduced in brief from the following aspects:
\begin{itemize}[]
    \item \textbf{Basic information.} Our guidelines learn from those of several widely used NER datasets to clarify how to write with correct format \citep{tjong-kim-sang-de-meulder-2003-introduction,xu2020cluener2020,Derczynski_2017_Results,Smith_2008_Overview,weischedel2011ontonotes}. To ensure that annotation guidelines fit entity labeling tasks, they are prepared by all annotators with the degree of Computer Science and some experienced experts give invaluable advice in confusing circumstances. More significantly, from an initial draft to mature version, our annotation guidelines go through lots of revisions and inspections during weekly meetings and finally become reliable enough \citep{Islamaj_2021_NLM-Chem}.
    \item \textbf{Main task.} The goal of our guidelines is illustrating whether tokens in each sentence should be tagged or not, and thus three fundamental considerations arise in annotation: whether the tokens are meaningful entities; which entity type the tokens belong to; how to determine entity boundary of the tokens. Moreover, our guidelines are required to explain entity categories and deliver a few instances for each type.
    \item \textbf{General rules.} To ensure consistent annotation, we establish clear directives, each supported by illustrative examples \citep{Islamaj_2021_NLM-Chem}. Several key principles govern our annotation process: First, generic agricultural terms such as ``crops", ``pests", and ``cereals" are excluded from entity annotation as they represent broad conceptual categories rather than specific entities. Second, taxonomic classifications (Order, Family, Genus) are consistently categorized as ``other organism" to maintain clear distinction from specific crops or livestock entities. Third, we adopt a unified approach for phenomenological entities - both regular weather events and meteorological disasters are classified under ``meteorological phenomenon", with the same principle applied to hydrological phenomena and hazards. Finally, agricultural areas without precise geographical specifications, such as ``soybean area" and ``Mulberry Garden", are excluded from annotation to maintain geographical precision in our dataset.
\end{itemize}

\subsection{Human annotation} 
\begin{figure}[t]
\centering
\includegraphics[width=\linewidth]{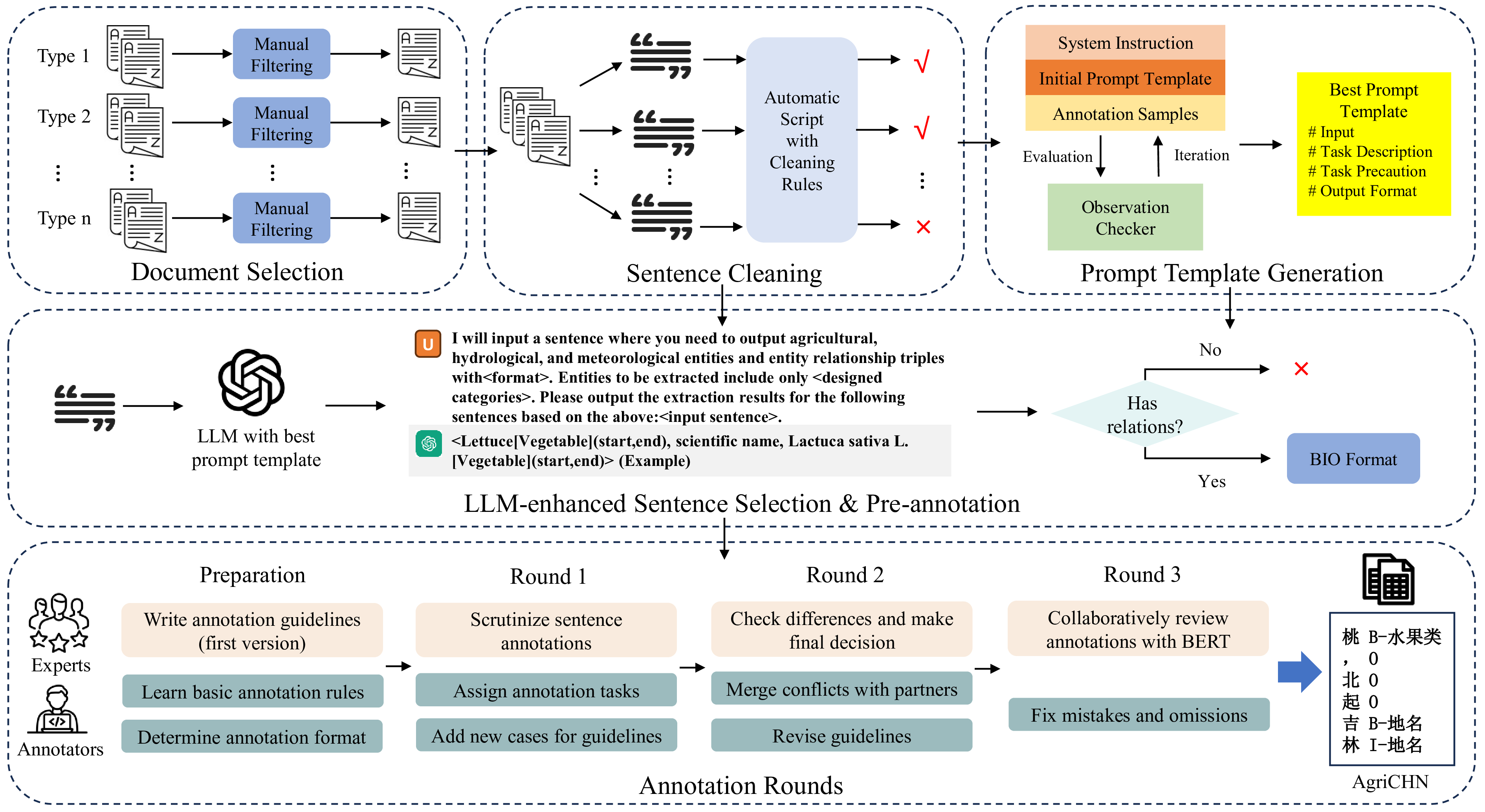}
\caption{Overall process of human annotation.}
\label{fig:Preparation-Annotation}
\end{figure}

Being desperately complex and time-consuming, human annotation is the most essential step in NER dataset construction. For annotators, they need to possess fluent proficiency in the target language, have basic knowledge of the field to which the dataset belongs, and be familiar with the specifications and requirements of annotation tasks \citep{yang-etal-2018-distantly}. Our annotation team is composed of 20 annotators skilled in entity identification and 5 specialists in agriculture. Before annotation task starts, all annotators receive a two-week training to understand relevant principles, carefully read the first version of annotation guidelines and provide suggestions for preliminary guideline improvement \citep{Islamaj_2021_NLM-Chem}. Then, annotators work on entity labeling for eight consecutive weeks. We design 3 rounds to iterate AgriCHN on the basis of the draft dataset produced from the step mentioned in \ref{sec:pre-annotation}. The overall process is shown in \hyperref[fig:Preparation-Annotation]{Fig. \ref*{fig:Preparation-Annotation}}.

In the first round, each annotator is assigned the same number of sentences to annotate independently. Meanwhile, each sentence is annotated by two annotators unaware of each other's identity according to task allocation. Then, their annotations for sentences are scrutinized by the expert group. If the annotation results fail to meet the established approval criteria, the corresponding annotators are required to relabel the sentences that do not meet the quality standards. 

In the second round, annotators begin to merge conflicts in annotations. Specifically, each pair of annotators responsible for the same sentence segments engage in discussions regarding their annotations to achieve consensus. If they still disagree on certain parts, an experienced expert check their differences and discrepancies and make the final decision.

In the third round, the expert group leverage BERT and take AgriCHN generated from the second round as the basic dataset to pre-train a NER model and predict labeling results by inputting sentences in this version of AgriCHN. Combing these results and expertise in agriculture, experts collaboratively review and revise annotations again to avoid possible mistakes and omissions. After this round, we get the final version of AgriCHN.

Throughout the annotation process, sentences in AgriCHN are annotated in batches, and each batch contains about 1000 sentences. Meanwhile, our annotations adopt the format named ``BIO'' and are stored as ``.txt'' files \citep{Dobbins_2022_Leaf}. Each line in the files includes a character, a NER tag (e.g., B-cereals) and a tab to separate them, and there is a new line between each sentence. Such simplified format makes the dataset easy to read and modify, allowing annotators and experts to pinpoint problems rapidly.

\subsection{Extra sentence replenishment}
After human annotation of agricultural articles, the AgriCHN dataset becomes accurate and extensive. However, it still needs improvement in entity distribution and sentence richness. Hence, we introduce sentences without candidate entities to implement data augmentation and sequences from other public available datasets such as CCIR2021 \citep{weischedel2011ontonotes}, MSRA \citep{levow2006third}, ResumeNER \citep{Zhang_2018_Chinese}, and WeiboNER \citep{Peng_2015_Named} for data balance.

\section{Dataset Statistics and Analysis}
\subsection{Overview}
\begin{figure}[t]
\centering
\includegraphics[width=\linewidth]{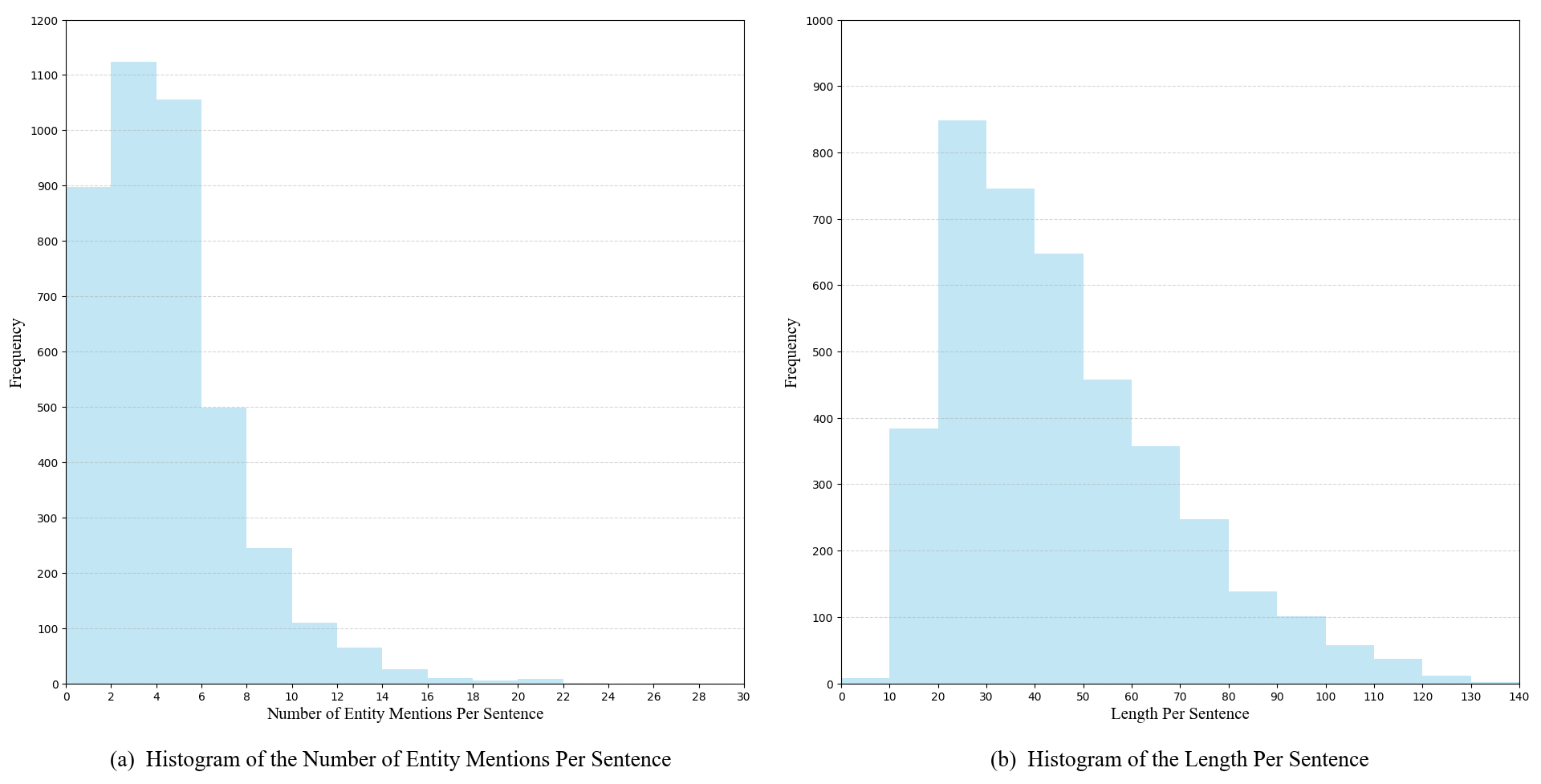}
\caption{Frequency histogram for number of total entity mentions and length per sentence.}
\label{fig:len_entity_sentence_distribution}
\end{figure}

AgriCHN is the first Chinese agricultural NER dataset with fine-grained entity types and covering cross-cutting fields including hydrology and meteorology. It contains 27 entity types, 4,040 unique sentences and 15,799 manual entity mention annotations in total, which are corresponded to 3 data sources: 550 hand-picked agricultural articles from authoritative sources, 728 extra sentences without entities obtained from paragraphs about smart agriculture, and 504 sentences extracted from various public NER datasets mentioned above to balance the number of entities. \hyperref[fig:len_entity_sentence_distribution]{Fig. \ref*{fig:len_entity_sentence_distribution}} depicts the distribution of entity mentions and length of sentences. The analysis reveals that, excluding the zero-shot evaluation sentences, AgriCHN demonstrates high information density with diverse entity mentions per sentence.
\subsection{Dataset split details}
For model development and evaluation, we split the dataset into training, development, and test sets in a ratio of 8:1:1. To ensure the reliability of model evaluation, we implement a multi-round manual adjustment process to achieve balanced distributions across all splits. Specifically, we examine and adjust the partitioning to maintain consistent patterns in both entity type frequency and sentence length distribution. This careful stratification effectively minimizes potential distribution biases that could impact model development and performance assessment. \hyperref[tab:train/dev/test]{Table \ref*{tab:train/dev/test}} presents comprehensive statistics for each data split, confirming the effectiveness of our balanced partitioning strategy.
\begin{table}[t]
\centering
\caption{\label{tab:train/dev/test}Distribution of agricultural annotations for train/dev/test set.}
\begin{tabular}{lcccc}
\toprule
Set & \makecell[l]{Characters} & \makecell[l]{Sentences} & \makecell[l]{Annotations} & \makecell[l]{Annotations Per Sentence} \\
\midrule
Train & 148,047 & 3,230 & 12,859 & 3.98 \\
Dev & 18,539 & 403 & 1,486 & 3.69 \\
Test & 18,217 & 407 & 1,454 & 3.57 \\
\bottomrule
\end{tabular}

\end{table}

\subsection{Inter-annotator agreement}
Inter-annotator agreement quantifies the consistency among different annotators performing the same NER task. Therefore, we conduct the agreement analysis on all double-annotated documents to validate the reliability of AgriCHN. In our evaluation framework, two annotations achieve exact agreement only when both annotators identify identical entity boundaries. Partial overlaps between entity mentions are treated as distinct annotations to maintain rigorous evaluation standards. We calculate Precision, Recall and F1-score as the measure of inter-annotator agreement. The statistical results reveal that: after annotators get used to the task and annotation guidelines are updated, the inter-annotator agreement scores of AgriCHN achieve 83.55\% in Precision, 78.36\% in Recall and 80.87\% in F1-score, indicating high degree of consistency despite of complex tasks. Additionally, we check annotation differences in detail and draw a conclusion: annotators tend to have disagreements mainly in entity boundaries, e.g., how to annotate modifiers in front of entity mentions.

\subsection{Entity \& sentence distribution among article types and parts}
\begin{figure}[t]
\centering
\includegraphics[width=0.9\linewidth]{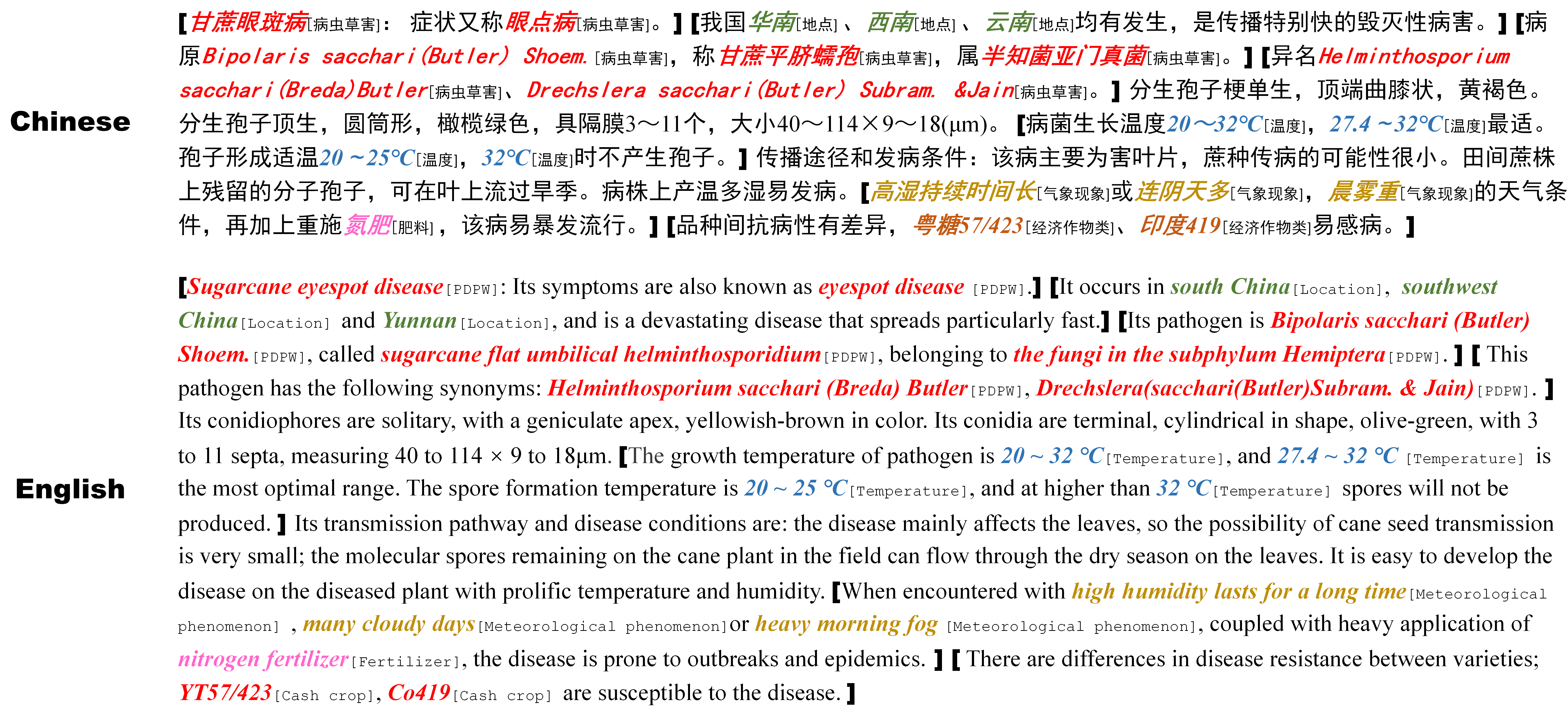}
\caption{Example of selected sentences in our proposed AgriCHN.}
\label{fig:example_selected_sentences}
\end{figure}

\hyperref[tab:Distribution of selected entities and sentences in different article type]{Table \ref*{tab:Distribution of selected and sentences in different article type}} shows the average and total number of entities and sentences with at least one entity in each article type. We find that articles about agricultural meteorology take the first place in average entity number in selected sentences and articles about soil moisture rank first in average selected sentence number in articles, which means that they contain enough mentions for corresponding entity types even if the article number is relatively small. Moreover, the distribution of selected sentences in articles follow special patterns. For example, in \hyperref[fig:example_selected_sentences]{Fig. \ref*{fig:example_selected_sentences}}, for an agricultural article about ``Sugarcane eye spot disease'', we mark all sentences containing golden agricultural entities with square brackets. Meanwhile, entities belonging to different categories are annotated in distinct colors. From annotation, the article mainly includes the following useful parts: synonym, distribution, pathogen, pathogen synonym, temperature, weather condition, corresponding varieties and prevention and control methods. In practical situations, these sentences are indeed highly valuable sections in agricultural articles. Thus we summarize that our preference in document and sentence selection process is beneficial for maintaining essential part of data sources.
\begin{table}[t]
\centering
\caption{\label{tab:Distribution of selected and sentences in different article type}Distribution of selected sentences in different article types.}
\resizebox{\linewidth}{!}{
\begin{tabular}{lcccccc}
\toprule
\makecell[l]{Article \\ Type} & 
\makecell[l]{Number of \\ Articles} & 
\makecell[l]{Number of Selected \\ Sentences} & 
\makecell[l]{Number of \\ Entities} & 
\makecell[l]{Number of Entities \\ Per Sentence} & 
\makecell[l]{Number of Selected \\ Sentences Per Article} \\
\midrule
Agricultural Knowledge & 200 & 1,099 & 5,023 & 4.57 & 5.49 \\
\makecell[l]{Plant Diseases, \\ Pests and Weeds} & 150 & 704 & 3,077 & 4.37 & 4.65 \\
Farming Guidance & 100 & 748 & 4,571 & 6.11 & 7.48 \\
Agricultural Machinery & 50 & 7 & 15 & 2.14 & 0.30 \\
Agricultural News & 30 & 77 & 293 & 3.81 & 2.57 \\
Soil Moisture & 10 & 129 & 704 & 5.46 & 12.90 \\
Agricultural Meteorology & 10 & 44 & 270 & 6.14 & 4.40 \\
\bottomrule
\end{tabular}
}

\end{table}

\subsection{Entity type distribution}
\begin{figure}[t]
\centering
\includegraphics[width=0.95\linewidth]{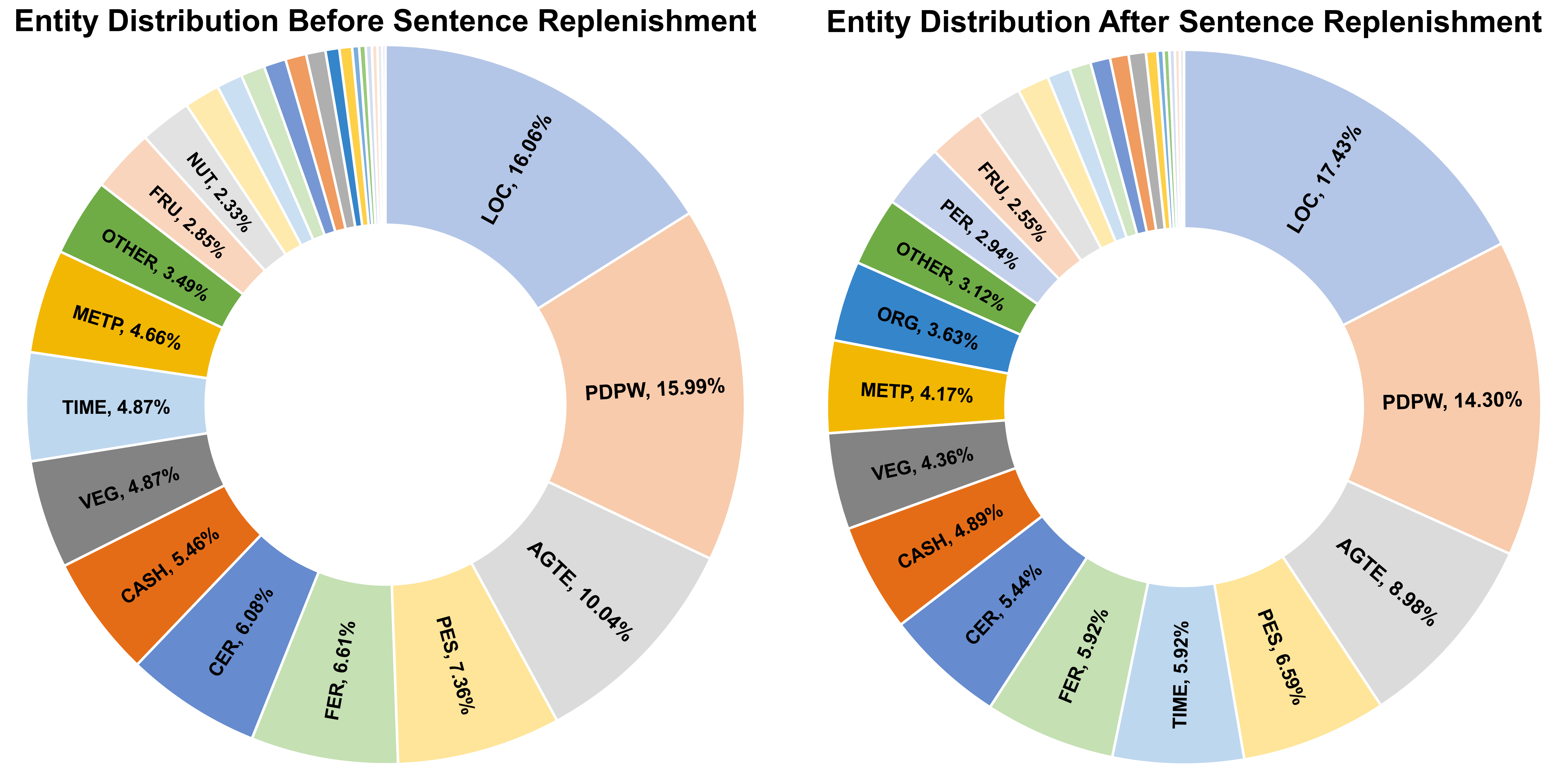}
\caption{Comparison of entity mentions for different entity types (AgriCHN \& AgriNER).}
\label{fig:entity_distribution_replenishment}
\end{figure}

We also analyze the entity type distribution in AgriCHN and address the issue of imbalanced distribution, which could potentially impact the performance of NER models. Our analysis reveal that the initial version of AgriCHN, derived solely from agricultural articles, contain notably few general entity types such as location, organization, person, and time. To address this limitation, we augment the dataset by incorporating 504 sentences from public datasets rich in these entity types. \hyperref[fig:entity_distribution_replenishment]{Fig. \ref*{fig:entity_distribution_replenishment}} illustrates the shift in their proportion, demonstrating a more balanced distribution.

\subsection{Comparison with other datasets}
\begin{figure}[t]
\centering
\includegraphics[width=\linewidth]{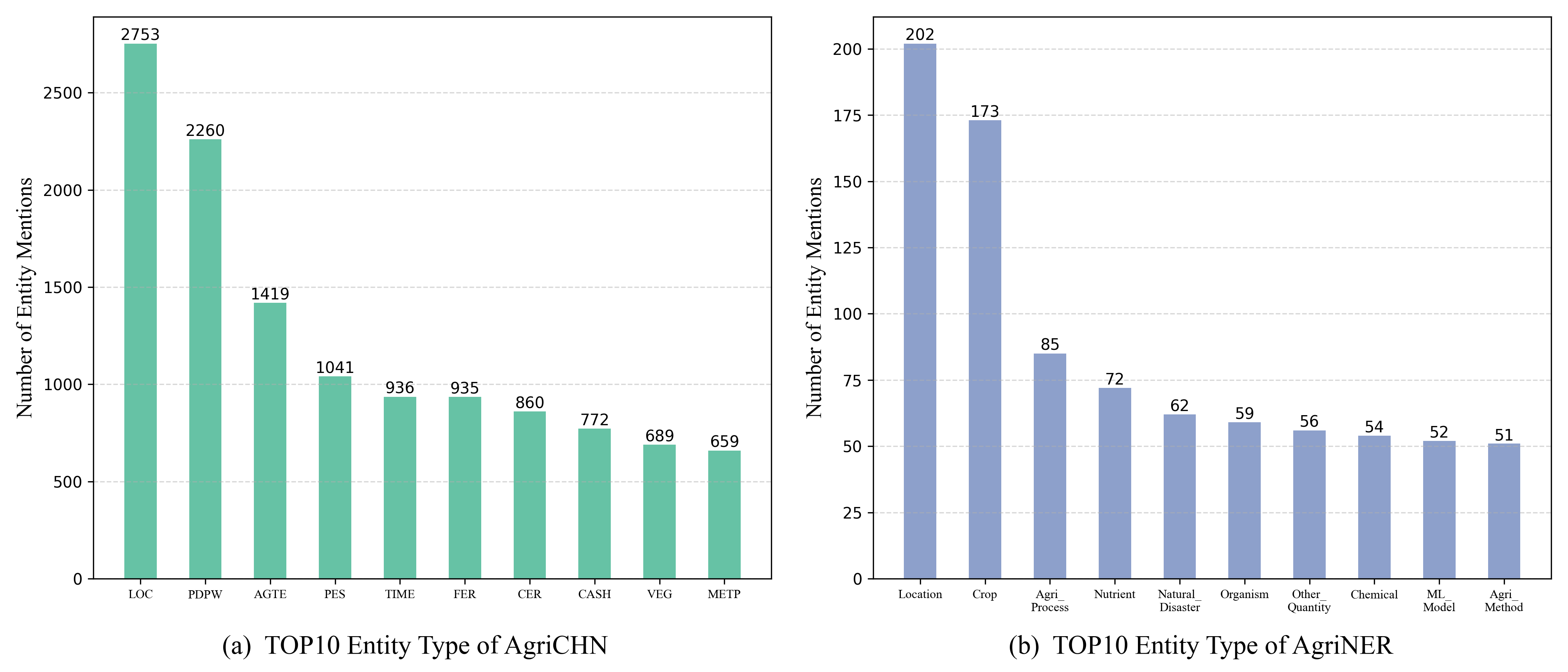}
\caption{Distribution of top-10 entity types in AgriCHN and AgriNER.}
\label{fig:entity_mentions_AgriCHN_AgriNER}
\end{figure}

To validate data quality, we compare the AgriCHN dataset with two kinds of datasets separately: similar Chinese agricultural NER datasets and other recent agricultural NER datasets. It is worth mentioning that unlike general NER, agricultural NER has no prominent and commonly used datasets, so we select recently published high-quality agricultural NER datasets for validation. \hyperref[tab:comparison with chinese sets]{Table \ref*{tab:comparison with chinese sets}} shows the comparison with several Chinese agricultural NER datasets, including WdpDs \citep{zhang2023awdpcner}, CCNEDP \citep{Wang_2022_Named}, KIWID \citep{zhang2022lexicon}, ApdCNER \citep{zhang2021chinese} and AgCNER \citep{guochinese2020}. Considering entity types and distributions, AgriCHN is not limited in common entity categories such as pests, diseases, crops, etc. It covers a wider variety of agricultural entities and becomes more useful for practical applications. Meanwhile, since Chinese datasets generally have a limited number of entity categories (fewer than 20), we find an English dataset named AgriNER\citep{de2023agriner} with a greater variety of entity types for a more comprehensive comparison. It comprises 36 fine-grained and self-explanatory entity categories, and also can be exploited to build a knowledge graph. As shown in \hyperref[fig:entity_mentions_AgriCHN_AgriNER]{Fig. \ref*{fig:entity_mentions_AgriCHN_AgriNER}}, AgriCHN has more agricultural entity mentions in most entity types than AgriNER. Our analysis yields that compared with AgriNER, AgriCHN not only has more sufficient examples for entity recognition but also optimizes entity type selection.

\begin{table}[t]
\centering
\caption{\label{tab:comparison with chinese sets} Chinese agricultural NER dataset comparison.}
\resizebox{\linewidth}{!}{
\begin{tabular}{lcccccc}
\toprule
Datasets & Entity Types & Entity Mentions & Dataset Size & Subdomain & Open Source \\
\midrule
AgriCHN & 27 & 15,799 & 4,040 sentences & Agriculture, Hydrology, and Meteorology & \checkmark \\
WdpDs \citep{zhang2023awdpcner} & 21 & 18,127 & \textasciitilde250,000 tokens & Wheat & $\times$ \\
CCNEDP \citep{Wang_2022_Named} & 3 & 7,806 & 1,161,820 tokens & Agriculture & \checkmark \\
KIWID \citep{zhang2022lexicon} & 6 & 17,809 & 12,477 sentences & Kiwifruit & $\times$ \\
ApdCNER \citep{zhang2021chinese} & 21 & 11,876 & 5,574 sentences & Apple & $\times$ \\
AgCNER \citep{guochinese2020} & 11 & 350,725 & 34,952 sentences & Agriculture & $\times$ \\
\bottomrule
\end{tabular}
}
\end{table}

\section{Experimental results}
\subsection{Settings}
Following previous works \citep{yang-etal-2018-distantly,Chang_2021_Chinese,Wang_2022_Named}, we define that the prediction is considered as a true positive only if all tokens are matched. Meanwhile, we employ BERT-base-chinese as the implementation of the BERT model, with a maximum sentence length of 256 tokens across all BERT-based models. For AgriCHN, we train on each NER model for 10 epochs. The batch size is set to be 16 and the AdamW optimizer \citep{loshchilov2017decoupled} with a learning rate of 3e-5 is used for model optimization. All models are trained on NVIDIA RTX 3090 GPU. Finally, we evaluate the model’s entity recognition performance using Precision, Recall, and F1-score as the performance metrics.
\subsection{Results on NER models}
We evaluate six representative NER models: BiLSTM-CRF \citep{Huang_2015_Bidirectional}, SWA \citep{Yan_2020_Bidirectional}, BERT \citep{Chang_2021_Chinese}, Chinese-BERT-wwm \citep{Cui_2021_Pre-Training}, RoBERTa \citep{liu2019roberta}, and MTNER \citep{tong2021multi}. For each NER model, we perform training and testing on the AgriCHN dataset ten times, and then average the results to obtain final promising benchmark. As shown in \hyperref[tab:Performance of NER models on AgriCHN. ]{Table \ref*{tab:Performance of NER models on AgriCHN. }}, MTNER has achieved optimal results among all models. However, its Precision, Recall and F1-score are still less than 0.86. Compared with other NER tasks (e.g., OntoNotes 5.0 \footnote{\href{https://paperswithcode.com/dataset/ontonotes-5-0}{https://paperswithcode.com/dataset/ontonotes-5-0}.}) which can easily achieve over 90\% accuracy, the AgriCHN poses greater challenges and warrants further in-depth research. 

To further validate dataset consistency, we conduct 10-fold cross-validation using the BERT model. The dataset is divided into ten equal subsets, with nine subsets used for training and one for validation in each iteration. This process is repeated ten times, rotating the validation subset to ensure comprehensive evaluation across the entire dataset \citep{malarkodi2016named}. Since the stability of the BERT model has been proven across various NER tasks, this validation approach is able to effectively assess the inherent variance in the AgriCHN corpus. The results show consistent performance with an average precision of 83.71\%, recall of 79.02\%, and F1-score of 81.30\%. Statistical analysis reveals a standard deviation of 0.97 for F1-scores across all folds, with the highest fold achieving 83.37\% and the lowest reaching 79.17\%. This small performance variance, with a range of only 4.20\%, demonstrates the dataset's robust quality and stability.
\begin{table}[t]
\centering
\caption{\label{tab:Performance of NER models on AgriCHN. } Performance of current neural NER models on AgriCHN. }
\resizebox{\linewidth}{!}{
\begin{tabular}{lcccccc}
\toprule
\multirow{2}{*}{Model} & \multicolumn{3}{c}{Development Set} & \multicolumn{3}{c}{Test Set} \\
\cmidrule(lr){2-4} \cmidrule(lr){5-7}
& Precision & Recall & F1-score & Precision & Recall & F1-score \\
\midrule
BiLSTM + CRF & 0.7736 & 0.7353 & 0.7539 & 0.7676 & 0.7608 & 0.7642 \\
SWA & 0.7711 & 0.7669 & 0.7690 & 0.7700 & 0.7705 & 0.7702 \\
BERT & 0.8400 & 0.7739 & 0.8056 & 0.8415 & 0.7639 & 0.8009 \\
Chinese-BERT-wwm & 0.8432 & 0.7853 & 0.8132 & 0.8449 & 0.7688 & 0.8050 \\
RoBERTa & 0.8448 & 0.7914 & 0.8172 & 0.8389 & 0.7846 & 0.8108 \\
MTNER & 0.8526 & 0.8103 & 0.8309 & 0.8498 & 0.8024 & 0.8254 \\
\bottomrule
\end{tabular}
}

\end{table}

\subsection{Results on entity types}
\begin{figure}[t]
\centering
\includegraphics[width=\linewidth]{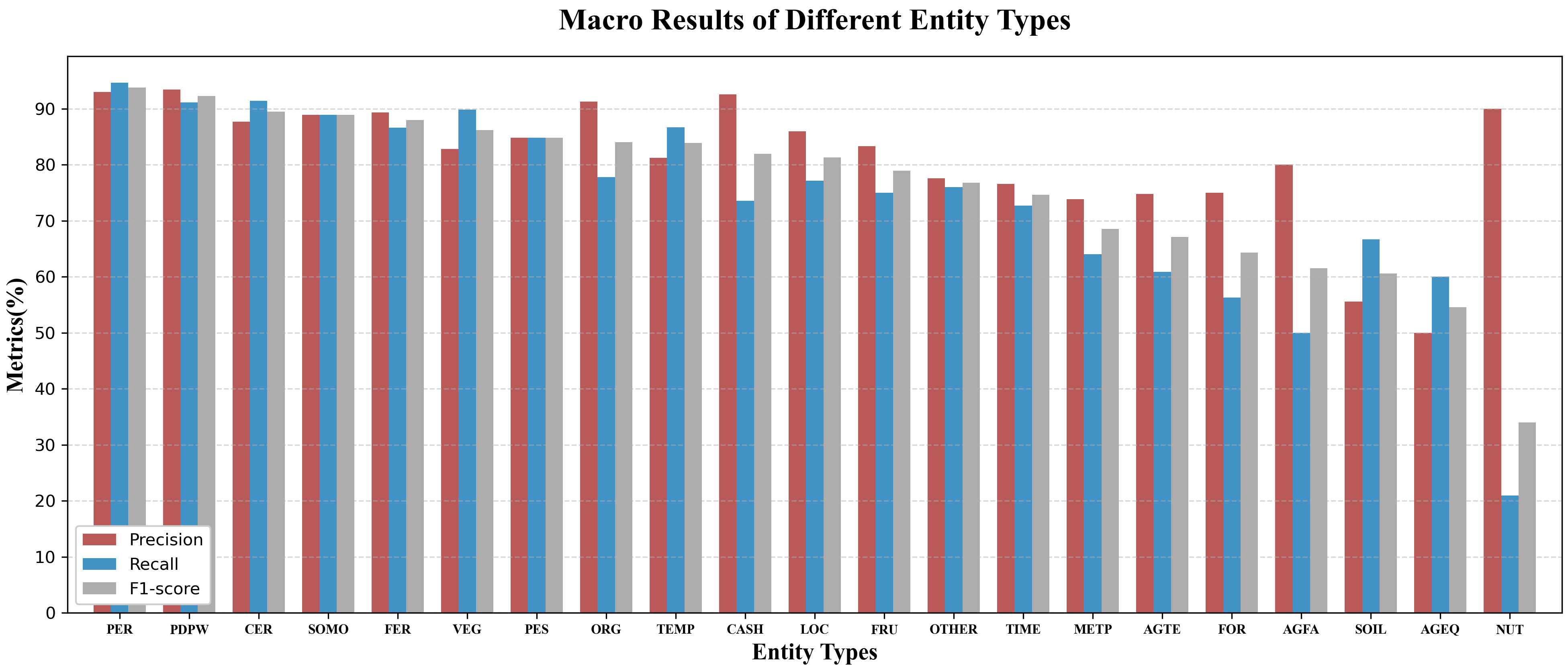}
\caption{Prediction result of Different Entity Types.}
\label{fig:P-R-F1-Entity-Type}
\end{figure}

We also perform a detailed analysis on the prediction performance across different entity types. As shown in \hyperref[fig:P-R-F1-Entity-Type]{Fig. \ref*{fig:P-R-F1-Entity-Type}}, among entity types with sufficient samples (more than 5 instances in the test set), the type ``PER" achieves the highest F1-score of 92.77\%, while the type ``NUT" exhibits the lowest F1-score of 33.96\%. This performance disparity is related to the inherent characteristics of different entity types. For instance, the type ``NUT" presents unique challenges in contextual interpretation. While standalone ``Nitrogen" is typically classified as a nutrient, ``Nitrogen fertilizer" should be labeled as fertilizer. This contextual ambiguity frequently occurs in agricultural texts and significantly affects the recall rate of nutrient entities. Moreover, our analysis reveals that prediction accuracy does not exhibit a linear correlation with sample size. A notable example is that the type ``AGTE" shows suboptimal recognition performance despite its substantial mentions in the dataset.

\subsection{Case study}
As shown in \hyperref[tab:Case_study]{Table \ref*{tab:Case_study}}, we provide a sample from AgriCHN to demonstrate the challenges faced by NER models. In this example, words that are underlined and have different fonts(chinese version: Kaiti; english version: italics) represent entity tokens, while words highlighted in color are the entity tokens identified by each NER model. By comparing the prediction results of the BERT and RoBERTa models with the golden entity mentions, we observe some deviations in their recognition of meteorological phenomena, water bodies, and nutrients. Specifically, the BERT model identifies the word ``freshwater" as the type ``location", which is a significant deviation. On the other hand, the RoBERTa model recognizes it as the type ``water body", which is more reasonable, but it still differs because this word is not considered as an entity mention. Moreover, it completely fail to recognize ``organic matter content". Meanwhile, neither model identifies the entity ``warm weather". Notably, all entities and entity categories mentioned are highly relevant to agricultural production, which indicates that entity recognition in AgriCHN is more challenging due to the diversity of the domain and the wide range of vocabulary involved.
\begin{table}[t]
\centering
\caption{\label{tab:Case_study} Case study results on the AgriCHN dataset.}
\begin{tabular}{lm{11.2cm}l}
\toprule
\multicolumn{1}{l}{Model} & Results \\ \hline
\multirow{1}{*}{BERT} & 
(EN) {\underline{\color{ForestGreen}{\textit{The phylum Cyanobacteria}}}} is widely distributed, primarily in {\color{ForestGreen}freshwater}, thrives in \underline{\textit{warm weather}} and waters with relatively high \underline{\textit{organic \color{ForestGreen}{matter} \color{Black}{content}}}, and possesses a strong ability to tolerate ultraviolet radiation and {\underline{\color{ForestGreen}{\textit{high temperatures}}}}... \\
\multirow{1}{*}{RoBERTa} & 
(EN) {\underline{\color{DodgerBlue}{\textit{The phylum Cyanobacteria}}}} is widely distributed, primarily in {\color{DodgerBlue}freshwater environments}, thrives in \underline{\textit{warm weather}} and waters with relatively high \underline{\textit{organic matter content}}, and possesses a strong ability to tolerate ultraviolet radiation and {\underline{\color{DodgerBlue}{\textit{high temperatures}}}}... \\
\bottomrule
\end{tabular}
\end{table}

\section{Discussion}
The AgriCHN is designed to be useful for various NLP tasks. Here we suggest two suitable usages of AgriCHN: training specific NER model in agricultural NER application; building agricultural knowledge graph.Currently, information extraction requires stable NER model and reliable datasets, especially in agricultural domain. By accurate entity identification, the application is able to summarize essential information in texts uploaded by users and thus assist their agricultural strategic decision-making. Additionally, the agricultural knowledge graph is a graph dedicated to representing and organizing knowledge in the agricultural domain, encompassing information, concepts, entities, relationships and attributes. It can help to integrate agricultural knowledge and make the knowledge easy to search and understand \citep{Chenglin_2018_Cn-MAKG}. Moreover, it is invaluable for improving the performance of search engines and personalised recommendation systems.

Though the AgriCHN corpus has been verified to be a robust and extensive resource, it does have several limitations that need to be noted. First, since the corpus is annotated by various annotators with multiple rounds, annotation disagreement problem still exists. Even if everyone has read the annotation guidelines in detail, annotators may make diverse judgments when faced with ambiguous labeling situations. As a result, two similar sequences might be annotated differently, which can potentially mislead NER model training. Additionally, AgriCHN needs improvement from the perspective of entity distribution. Some entity types including disease, water body, livestock, and hydrological phenomenon do not have enough samples in the development set and test set, and such imbalance cause lower performance in related entity identification. This can be derived to document and sentence selection to some extent, because in order to ensure the informativeness of agricultural texts, we select an abundance of articles about agricultural knowledge. These sentences contain a large number of entities belonging to the type ``PDPW" and types about crops rather than those uncommon entity types. Last, the size of AgriCHN is limited compared with similar agricultural NER datasets. In future work, we will select more agricultural articles and invite extra experts in entity recognition and agriculture domain to build a better version of AgriCHN.

\section{Conclusion}
In this paper, we introduce AgriCHN, a meticulously curated dataset designed to tackle the challenge of the limited availability of extensive and dependable corpora for Chinese agricultural NER. Annotated by 20 experienced annotators and reviewed by 5 experts in the agriculture domain with the help of LLM, the AgriCHN corpus experiences 3 rounds of iterative annotation and manual correction to attain an exceedingly high level of data quality. Meanwhile, it is the first Chinese agricultural NER dataset which covers entities about meteorology and hydrology. In terms of statistical data, the AgriCHN contains 27 fine-grained entity types, 4,040 sentences, and 15,799 annotated agricultural mentions; corresponding to 550 informative agricultural articles, 728 sentences about smart agriculture without entity mentions and 504 sentences extracted from publicly available datasets. To validate its availability, we perform the data characteristic analysis and conduct baseline experiments using six NER models. Extensive empirical results show that the AgriCHN corpus has notably refined entity classification criteria. Consequently, the entity recognition task using AgriCHN proves to be significantly more challenging for mainstream NER models when compared with other Chinese agricultural NER datasets.

\section{Acknowledgements}
The research work is supported by the National Key Research and Development Program of China under Grant Nos. 2024YFF0729003, 2022YFD\\1600602, the Reform and Development Project of Beijing Academy of Agricultural and Forestry Sciences, the National Natural Science Foundation of China under Grant Nos. 62176014, 62276015, the Fundamental Research Funds for the Central Universities.

\section*{Declaration of generative AI and AI-assisted technologies in the writing process}
During the preparation of this work the authors used ChatGPT in order to improve language and readability. After using this tool, the authors reviewed and edited the content as needed and take full responsibility for the content of the publication.

\bibliographystyle{elsarticle-harv}
\bibliography{references}
\end{document}